\documentclass[lettersize,journal]{IEEEtran}
\IEEEoverridecommandlockouts                              
\usepackage{cite}
\usepackage{booktabs}
\usepackage{makecell}
\usepackage{algorithm}
\usepackage{algpseudocode}
\usepackage{amsmath}
\usepackage{amsfonts}
\usepackage{amssymb}  
\usepackage[utf8]{inputenc}
\usepackage[english]{babel}
\usepackage[T1]{fontenc}
\usepackage{url}
\usepackage{soul}
\usepackage{color}
\usepackage{graphicx}
\usepackage{xcolor}
\usepackage[export]{adjustbox}
\usepackage{subcaption}

\newcommand{\ours}{\emph{BETR-GUI}}
\begin{document}
\bstctlcite{IEEEexample:BSTcontrol}


\renewcommand\theadfont{\bfseries}
\renewcommand\theadgape{\Gape[4pt]}

\author{Jonathan Styrud\IEEEauthorrefmark{1}\IEEEauthorrefmark{2}, Matteo Iovino\IEEEauthorrefmark{3}\IEEEauthorrefmark{4}, Rebecca Stower\IEEEauthorrefmark{5},  Mart Kartašev\IEEEauthorrefmark{2}, \\ Mikael Norrlöf\IEEEauthorrefmark{1}, Mårten Björkman,\IEEEauthorrefmark{2} and Christian Smith\IEEEauthorrefmark{2}
\thanks{This project is supported by the Wallenberg AI, Autonomous Systems, and Software Program (WASP) funded by the Knut and Alice Wallenberg Foundation. The authors gratefully acknowledge this support.}
\thanks{\IEEEauthorrefmark{1}ABB Robotics, Västerås, Sweden}
\thanks{\IEEEauthorrefmark{2}Department of Robotics, Perception and Learning, Royal Institute of Technology (KTH), Stockholm, Sweden}
\thanks{\IEEEauthorrefmark{3}Mobile Robotics Lab, ETH Zürich, Zürich, Switzerland}
\thanks{\IEEEauthorrefmark{4}ABB Corporate Research, Västerås, Sweden}
\thanks{\IEEEauthorrefmark{5}Design and User Experience, Ericsson, Stockholm, Sweden}}

\title{\LARGE Design and Evaluation of an Assisted Programming Interface for Behavior Trees in Robotics}
\maketitle

\begin{abstract}
The possibility to create reactive robot programs faster without the need for extensively trained programmers is becoming increasingly important. So far, it has not been explored how various techniques for creating Behavior Tree~(BT) program representations could be combined with complete graphical user interfaces~(GUIs) to allow a human user to validate and edit trees suggested by automated methods. In this paper, we introduce BEhavior TRee GUI (\ours{})  for creating BTs with the help of an AI assistant that combines methods using large language models, planning, genetic programming, and Bayesian optimization with a drag-and-drop editor. A user study with 60 participants shows that by combining different assistive methods, \ours{} enables users to perform better at solving the robot programming tasks. The results also show that humans using the full variant of \ours{} perform better than the AI assistant running on its own.
\end{abstract}

\def\abstractname{Note to Practitioners}
\begin{abstract}
Rapid development of reactive robot programs without requiring extensively trained programmers is becoming increasingly important in the robotics industry, especially as more robots are working in uncontrolled environments. The results from our user study indicate that a human working together with an AI assistant, when using a graphical user interface for robot programming with behavior trees, could significantly improve task performance over a human programmer or AI working on their own. We further suggest that the AI assistant should be based on a combination of multiple different algorithms that complement each other, such as planning, machine learning, and large language models.
\end{abstract}

\section{Introduction}
Robots have historically operated in controlled environments where they are capable of solving complex tasks with high reliability and precision, often repeating the same task for years. Today, however, robots are increasingly working in unpredictable environments~\cite{worldrobotics}, with mobile robots or workspaces shared with humans. Furthermore, as smaller businesses adopt robot automation, the trends are towards ever smaller batches of products or tasks and thereby more frequent updates of the robot programs. As a consequence, the possibility to create robot programs faster without the need for extensively trained programmers and for those programs to be reactive to their environment is becoming increasingly important. Another decisive factor is that the program should be transparent and readable to enable analysis, editing, and validation.
A policy representation that fulfills the aforementioned requirements and is steadily growing in popularity is behavior trees (BTs)~\cite{iovino_survey_2022, colledanchise_behavior_2018}. Related to this, research on facilitating the creation of BTs using various techniques is very active~\cite{iovino_survey_2022, colledanchise_synthesis_2017, colledanchise_learning_2019, colledanchise_towards_2019, rovida_extended_2017, iovino_learning_2021,  styrud_combining_2022, llmbt, llmobtea, styrud2025automatic} and multiple Graphical User Interfaces (GUIs) have been developed to facilitate manual design~\cite{groot2_webpage, becroft2011aipaint, shoulson_parameterizing_2011, guerin2015framework, paxton_costar_2017, paxton_evaluating_2018}. However, to our knowledge, it is so far unexplored how the various BT generation techniques could be combined under complete GUIs to let a human user validate and edit BTs suggested by these automated methods. In this paper, we introduce BEhavior TRee GUI (\ours{}), a GUI for creating BTs with the help of an AI assistant that combines several previous methods. Some recent results have indicated that software developers could, surprisingly, perform worse with AI assistants than without~\cite{becker2025measuring}. Nevertheless, our user study performed with \ours{} indicates that users perform significantly better using \ours{} with the AI assistant enabled.

The main contributions of this paper are:
\begin{itemize}
    \item A novel user interface, \ours{}, for creating BTs with an AI assistant that combines methods using large language models, planning, genetic programming, and Bayesian optimization with a drag-and-drop editor for humans and validation in a simulated environment.
    \item An extensive user study with 60 participants showing that for the tasks in the experiments, \ours{} combining the assistive methods enables users to perform better at solving the tasks than some ablated versions of the assistant or without the assistant at all. The results also show that humans using the full variant of \ours{} perform better than the AI assistant running on its own.
\end{itemize}

\section{Background and related work}

\subsection{Behavior Trees}
\begin{figure*}[hbtp]
\includegraphics[width=1.0\textwidth]{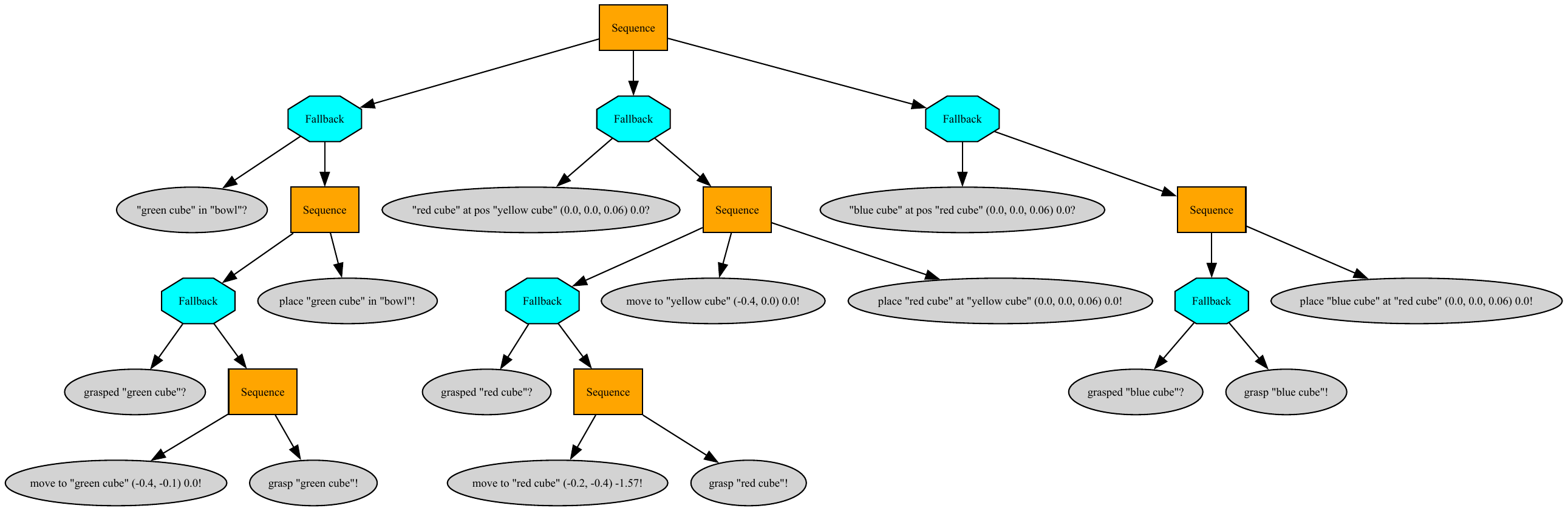}
\caption{Example of a non-optimal behavior tree for solving the \emph{Cubes and bowl} task. The different node types are described in Section~\ref{section:task_design}.}
\vspace{-.0cm}
\label{fig:example_bt}
\end{figure*}

Behavior Trees (BTs) is an architecture for reactive policies that originated in the computer gaming industry but has been increasingly used for robotics~\cite{colledanchise_behavior_2018, iovino_survey_2022}. An example BT for a robotics manipulation task can be seen in Figure~\ref{fig:example_bt}. The recent success of BTs can be attributed to their main advantages: readability, modularity, and ingrained support for task hierarchy, action sequencing, and reactivity. Readability enables manual and automated editing, analysis, and validation~\cite{colledanchise_how_2017}. These are strong advantages compared to black box architectures like neural networks. BTs are known to improve on other architectures as well, such as finite state machines, in terms of modularity and reactivity~\cite{iovino2025comparison,colledanchise_how_2017,biggar_modularity_2022}. In fact, BTs have been shown to be optimally modular~\cite{biggar2022modularity}.
\par Functionally, a BT is a directed tree where a tick signal propagates from the root node down to the leaves at some given frequency. Nodes are executed only when they receive the tick and return one of the three states \emph{Success}, \emph{Failure}, and \emph{Running}. The branching non-leaf nodes are called \emph{control flow nodes} or just \emph{control nodes}. The most common control nodes are \emph{Sequence}, which ticks children sequentially from left to right, returning \emph{Success} once all succeed or \emph{Failure} if any child fails, and \emph{Fallback}, (or \emph{Selector}) which also ticks from left to right but returns \emph{Success} when any child succeeds or \emph{Failure} only when all children fail. Leaves are called \emph{execution nodes} or \emph{behaviors} and are typically separated into \emph{Actions}(``!'') and \emph{Conditions}(``?''). Conditions denote quick status checks or sensory readings, always returning \emph{Success} or \emph{Failure} directly. Actions denote robot skills that can take more than one tick to complete and return \emph{Running} while execution is ongoing. For more details, we refer to~\cite{colledanchise_behavior_2018}.
\par While BTs have multiple advantages, they still take time and effort to design, and more efficient methods to create BTs have been given substantial research interest in recent years~\cite{iovino_survey_2022}, using learning methods~\cite{colledanchise_learning_2019, iovino_learning_2021, iovino2023framework, mayr22priors}, analytical planners~\cite{tumova_maximally_2014, colledanchise_synthesis_2017, colledanchise_towards_2019, holzl_reasoning_2015, rovida_extended_2017}, improved user interfaces~\cite{gustavsson_combining_2022, gugliermo2023learning, iovino2022interactive}, large language models~\cite{izzo2024btgenbot, mower2024ros} or combinations of the various methods~\cite{mayr2022combining, mayr2022skill, styrud_combining_2022, styrud2024bebop, llmbt, llmobtea, chen2024efficient, styrud2025automatic}.

\subsection{Planning}
Automated planning algorithms~\cite{ghallab_automated_2016} have been extensively used to facilitate the creation of BTs. In this paper we use an adaptation of the Planning Domain Definition Language (PDDL)-style planner from~\cite{colledanchise_towards_2019}, creating back-chained BTs. We build on the latest adaptation of the same planner from~\cite{styrud2025automatic}. The foremost reason for this choice of planner is its simplicity, but there are other more advanced planners for BTs to consider as well, for example, Linear Temporal Logic (LTL)~\cite{tumova_maximally_2014,colledanchise_synthesis_2017} 
and Hierarchical-Task-Network (HTN) planning~\cite{holzl_reasoning_2015, rovida_extended_2017}. We refer to Sections 4.1.2 and 2.4 in~\cite{iovino_survey_2022} for a more exhaustive list.
\par With sufficient time and knowledge, formal planners can, in theory, solve all common robotics task planning problems. In reality, the engineering effort to formalize that knowledge is substantial, and there will always be applications where some information is missing. This has been addressed with some success through the use of large language models~\cite{llmbt, llmobtea, chen2024efficient, styrud2025automatic, ahmad2025vision, ahmad2025unified}.

\subsection{Genetic Programming}
\label{sec:gp}
Genetic Programming (GP) is an optimization algorithm that evolves populations of programs in the form of trees~\cite{koza_genetic_1992,sloss_2019_2020}. In short, the populations generate offspring through \emph{crossover} and \emph{mutation}, after which a selection algorithm decides which individuals to keep based on some fitness function that evaluates and scores each individual. Some common examples of fitness functions are \emph{elitism} (select $n$ highest ranking), \emph{tournament selection} (individuals compared pair-wise and the lowest scoring is eliminated), and \emph{rank selection} (random selection where an individual's probability of survival is proportional to its rank within the population). There are many variations of GP, for example, where a grammar is defined and programs are generated from simple lists of integers (grammatical evolution), or where the genotype is represented as fixed-length strings (genetic algorithms).
\par GP has also been successfully used to evolve BTs specifically. Crucially, several studies have shown that GP can find solutions that outperform manual designs~\cite{hutchison_evolving_2010,scheper_behavior_2015,jones_evolving_2018,perez_evolving_2011,colledanchise_learning_2019}. As BTs originated in the gaming industry, video game benchmarks~\cite{perez_evolving_2011,nicolau_evolutionary_2017,colledanchise_learning_2019, hutchison_evolving_2010, mcclarron_effect_2016} are still more common than robotics~\cite{scheper_behavior_2015, jones_evolving_2018, iovino_learning_2021, styrud_combining_2022}. For the GP algorithm to be efficient, it is typically necessary to employ various constraints to minimize the search space. Using more constraints could speed up the learning but may instead not generalize to some tasks. Some constraints used in~\ours~were inspired by~\cite{mcclarron_effect_2016, yang2025learning} with some modifications described in the upcoming Section~\ref{sec:assistant}. For example, allowing any control node type as the root.

\subsection{Bayesian Optimization}
Bayesian optimization (BO) is a method to efficiently solve variable optimization problems while limiting the number of expensive evaluations. It has shown great performance in various applications, such as robotics~\cite{calandra2016bayesian, mayr2022skill, rai2018bayesian, styrud2024bebop}, hyperparameter tuning~\cite{klein2017fast, kandasamy2018neural, ru2021interpretable}, and material design~\cite{frazier2015bayesian, packwood2017bayesian, hughes2021tuning}.
\par BO iteratively selects new values to evaluate, while handling the  trade-off between exploration and exploitation. It uses an internal surrogate model of the unknown true objective function that it learns as it gathers more data. The two most common models are Gaussian processes~\cite{williams2006gaussian} for their ability to quantify uncertainty on top of providing accurate predictions, and random forests~\cite{lindauer2022smac3, shahriari2015taking} for their versatility, for example, better representation of non-smooth objective functions, and scalability to more samples. We refer to~\cite{frazier2018tutorial} for a more detailed introduction to BO.

\subsection{LLMs}
The widespread adoption of Large Language Models (LLMs)~\cite{radford2019language, brown2020language} in recent years can hardly have evaded anyone, and the field of robotics is not excluded~\cite{wang2024survey, saycan}. Various methods have been proposed that interpret the users' natural language instructions to create robot programs~\cite{radford2019language, brown2020language, wang2024survey} or specifically to create XML files defining reactive BT policies~\cite{izzo2024btgenbot, mower2024ros, lykov2023llm}. Creating a policy instead of a sequential plan that would need to be continuously re-evaluated minimizes the number of calls to the LLM, especially during execution, saving time and cost. It also makes the policy transparent and verifiable before execution, which can be a hard requirement for many applications. 
\par LLMs are, however, known to struggle with complex tasks requiring long-horizon planning~\cite{shojaee2025illusion}. To overcome this limitation, some methods are based on the LLMs producing PDDL-format~\cite{aeronautiques1998pddl} problem descriptions from natural language instructions~\cite{liu2023llm+}, which enables the use of PDDL planners to solve the tasks. Combining LLMs and planners to create BTs has also received some attention lately. LLM-BT~\cite{llmbt}, uses an extensive hard-coded parser to translate the LLM's responses into an initial BT that is then expanded with a planner. The similar method LLM-OBTEA~\cite{llmobtea} uses stricter prompting to obtain goal conditions directly with less parsing and introduces an additional step called reflective feedback, which iteratively prompts the LLM to solve syntactical errors. It's extended in HOBTEA~\cite{chen2024efficient} where an LLM suggests prioritized areas of the search space to speed up the planning algorithm. Finally, BETR-XP-LLM\cite{styrud2025automatic} showed that an improved prompt and LLM model eliminate the need for reflective feedback, giving better results in fewer LLM calls. BETR-XP-LLM also introduced a method to use the LLM to resolve errors outside the planner's capabilities during planning and execution. The later work of~\cite{merino2025behavior} also includes a failure handler as well as a module to ask the user for clarifications. \cite{ahmad2025vision} and~\cite{ahmad2025unified} expanded on~\cite{styrud2025automatic} by replacing the LLM with a vision language model with access to a scene graph and execution history, and adding the capability to also add missing skill templates rather than only adding missing preconditions to existing skills.

\subsection{Combinations}
Multiple previous works attempt to combine various methods into composite systems where the methods complement each other so that the resulting systems perform better than their parts. With a simplified grouping of methods, it can roughly be said that manual design through user interfaces tends to give the user full control but requires substantial amounts of time and skill. Automated planning tends to give very predictable and structured solutions and is often very fast, but can scale poorly on very complex problems and depends heavily on having structured knowledge in advance. Finally, learning methods can be slower by many orders of magnitude for simpler tasks but excel on complex problems and need no previous knowledge. Their main drawback is that they tend to give solutions that are difficult to verify, validate, edit, and debug. Some notable examples of work on composite systems for creating BTs include planning with GP~\cite{styrud_combining_2022}, as well as with other learning methods~\cite{mayr2022combining, mayr2022skill}, learning from demonstration and GP~\cite{iovino2023framework}, LLMs and planning~\cite{llmbt, llmobtea, chen2024efficient, styrud2025automatic}, and LLMs and GP~\cite{kobilov2025automatic}. In this paper, we introduce a comprehensive composite system that integrates multiple previous works, including planning, genetic programming, Bayesian optimization, LLMs, and a GUI for manual editing, and perform an ablative user study to determine the contributions of the different components.

\section{Interface Design}
\label{sec:interface_design}
\begin{figure*}[hbtp]
\hspace*{-0.5cm}
\includegraphics[width=1.4\textwidth]{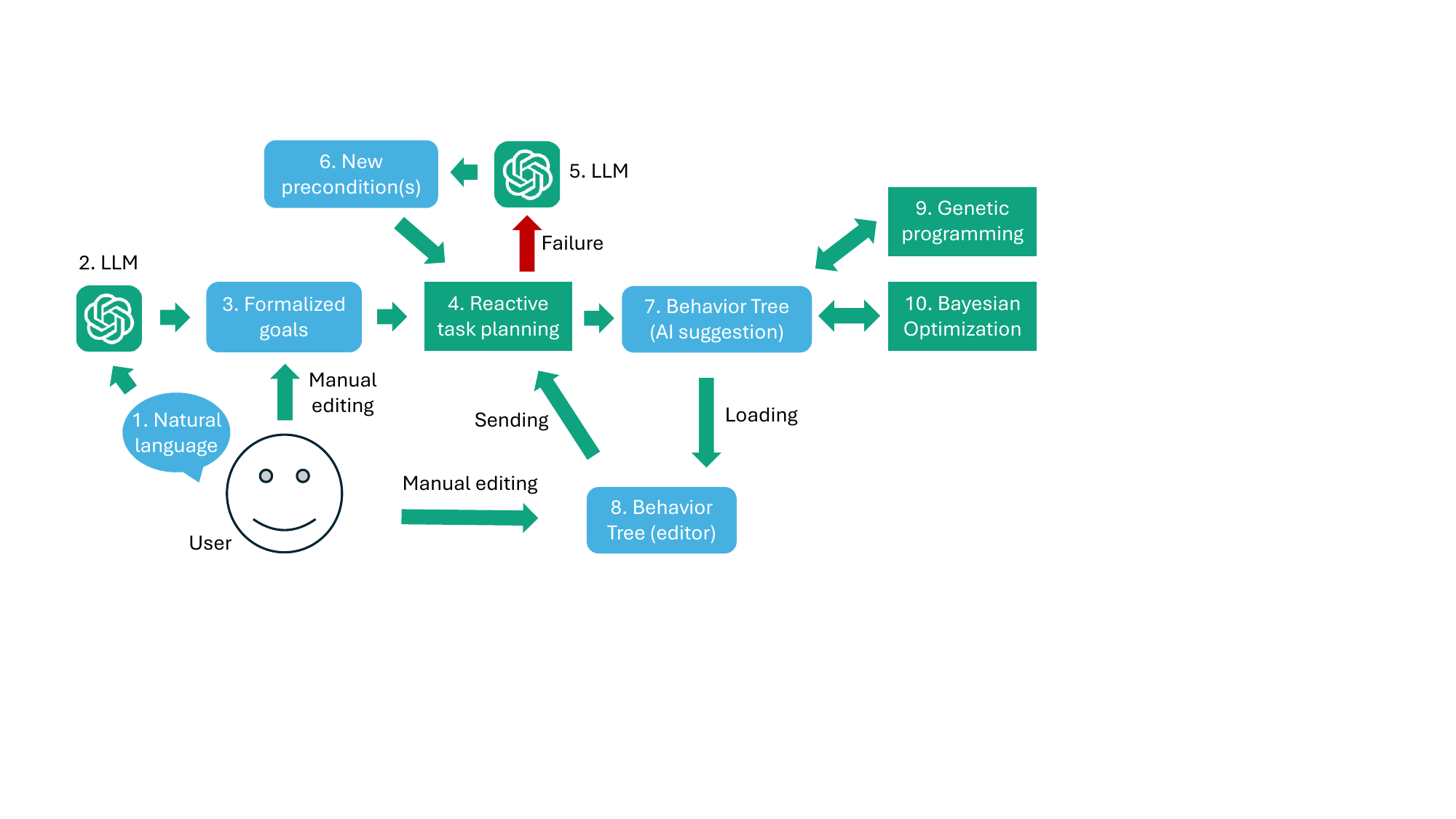}
\vspace{-3.6cm}
\caption{Graphic representation showing an overview of the algorithms of \ours~and their connections. Green boxes denote algorithms, and blue boxes denote data.}
\vspace{-.0cm}
\label{fig:method}
\end{figure*}

In this section, we describe the design and implementation of the graphical user interface used for the experiments. Overall, the idea was to construct a minimal interface that combines various methods for creating behavior trees. An overview is shown in Figure~\ref{fig:method}. The LLM model used in all experiments was GPT-4, but there are no special adaptations specifically to this LLM variant. The code for the interface was written in Python, using PyQt5 for graphics, and is publicly available on GitHub\footnote{https://github.com/jstyrud/BETR-GUI}. The interface was created in six different variants as described in Section~\ref{sec:variants}.

\subsection{Instructions Tab}
\begin{figure*}[hbtp]
    \vspace{0.0cm}
\hspace*{-0.15cm}
\includegraphics[width=1.0\textwidth]{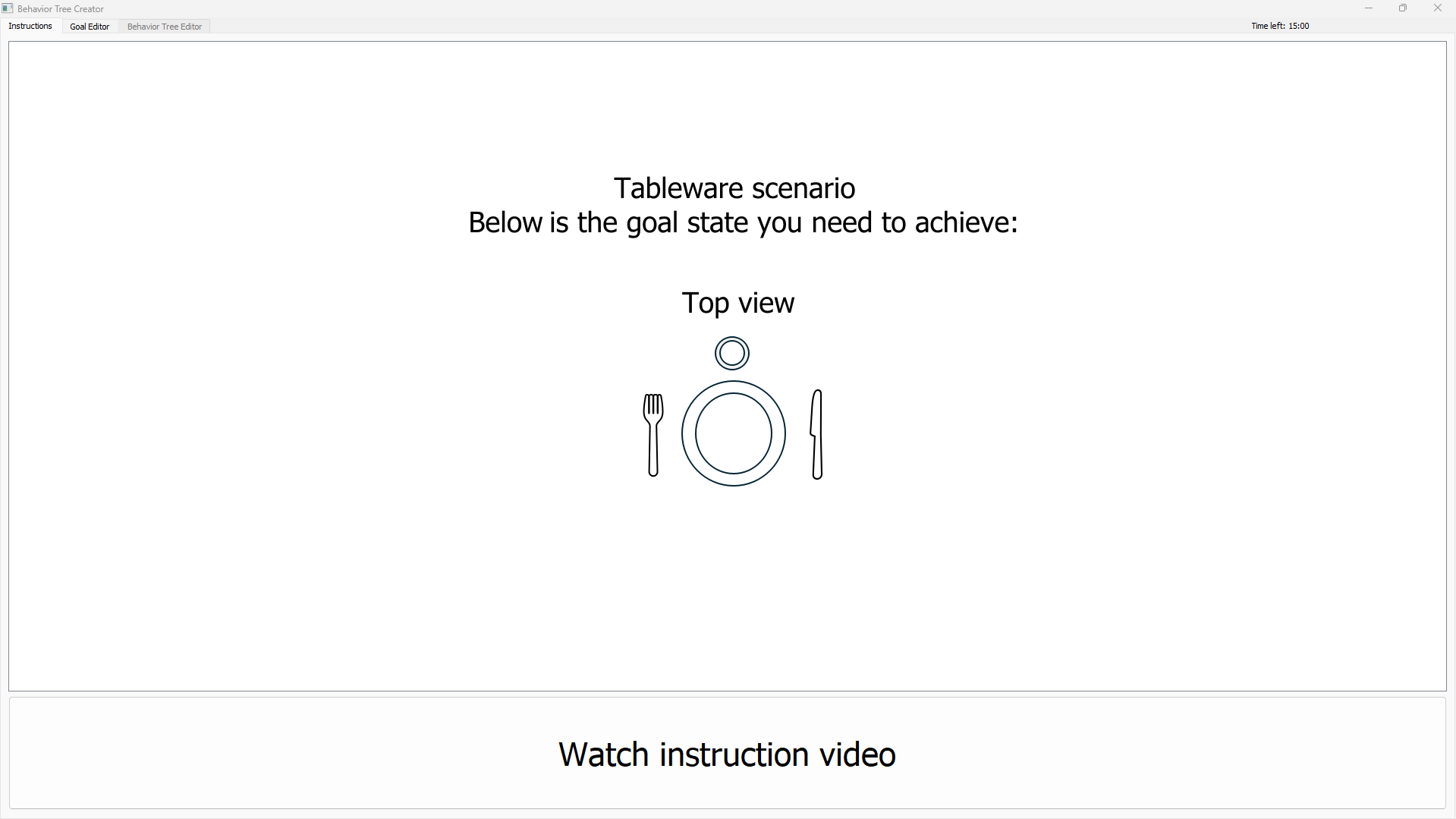}

\vspace{-0.0cm}
\caption{Screenshot of the Instructions tab of the GUI with the \emph{Tableware} scenario description.}
\vspace{-.0cm}
\label{fig:tab1}
\end{figure*}
The first tab of the interface shown to the user is the Instructions Tab. An example is shown in Figure~\ref{fig:tab1}. This tab shows a graphical description of the present task as well as a link to the instruction video.
\subsection{Goal Editor Tab}
\begin{figure*}[hbtp]
\vspace{0.0cm}
\hspace*{-0.15cm}
\includegraphics[width=1.0\textwidth]{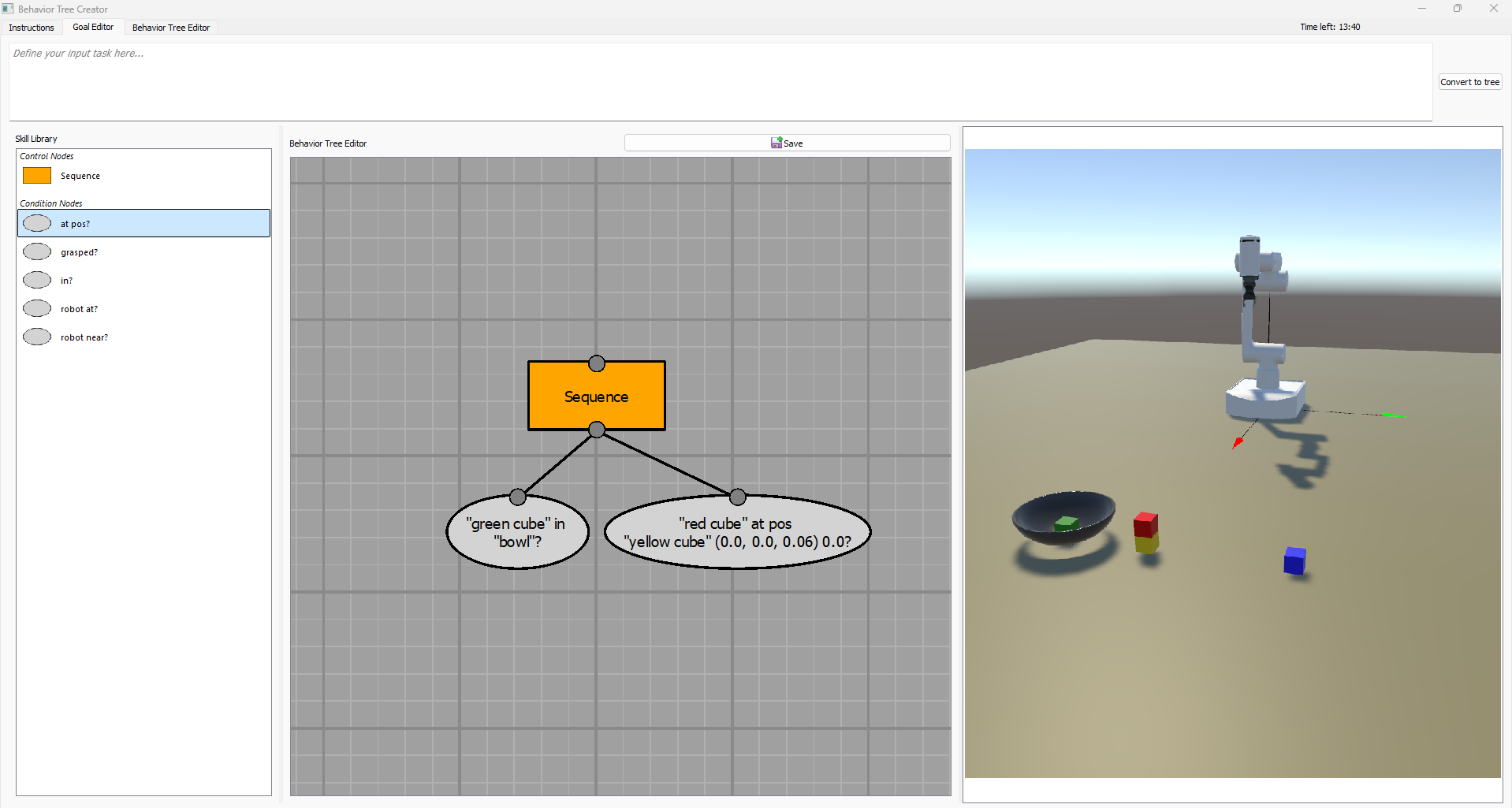}

\vspace{-0.0cm}
\caption{Screenshot of the Goal Editor tab of the GUI with the \emph{Cubes and bowl} task with the goal conditions currently solving part of the task, as the goal condition of the blue cube is not yet correctly defined.}
\vspace{-.0cm}
\label{fig:tab2}
\end{figure*}
The second tab is the Goal Editor Tab, where the user defines the goal state in a format that the robot can understand. An example is shown in Figure~\ref{fig:tab2}. The experiment timer starts counting down when the user first enters this tab, ensuring that all users get the same amount of time for each task. To the left, the user has a list of available condition nodes that can be used to define the goal conditions as well as the available control nodes, which in this case is only the \emph{Sequence} node. The reason for omitting \emph{Fallback} at this stage is purely to keep the user experience simple enough for novice users. After the user has dragged them into the editor widget, the user can define the parameters by double-clicking the node to open a separate window for parameter editing. After saving, the user can see a rendered image of the goal state to the right to confirm the goal conditions visually. 
\par At the top of the Goal Editor Tab, there is a text field where the user can describe the goal in natural language and have it converted automatically into a tree with condition nodes. This text field is available in all variants except MANUAL\_ONLY and NO\_LLM. This functionality uses the same method as in~\cite{styrud2025automatic} with a slightly updated prompt and some added error handling. After having the text converted, the user is free to edit the suggested tree or to update the prompt to iteratively improve.
\subsection{Behavior Tree Editor Tab}
\begin{figure*}[hbtp]
\vspace{0.0cm}
\hspace*{-0.15cm}
\includegraphics[width=1.0\textwidth]{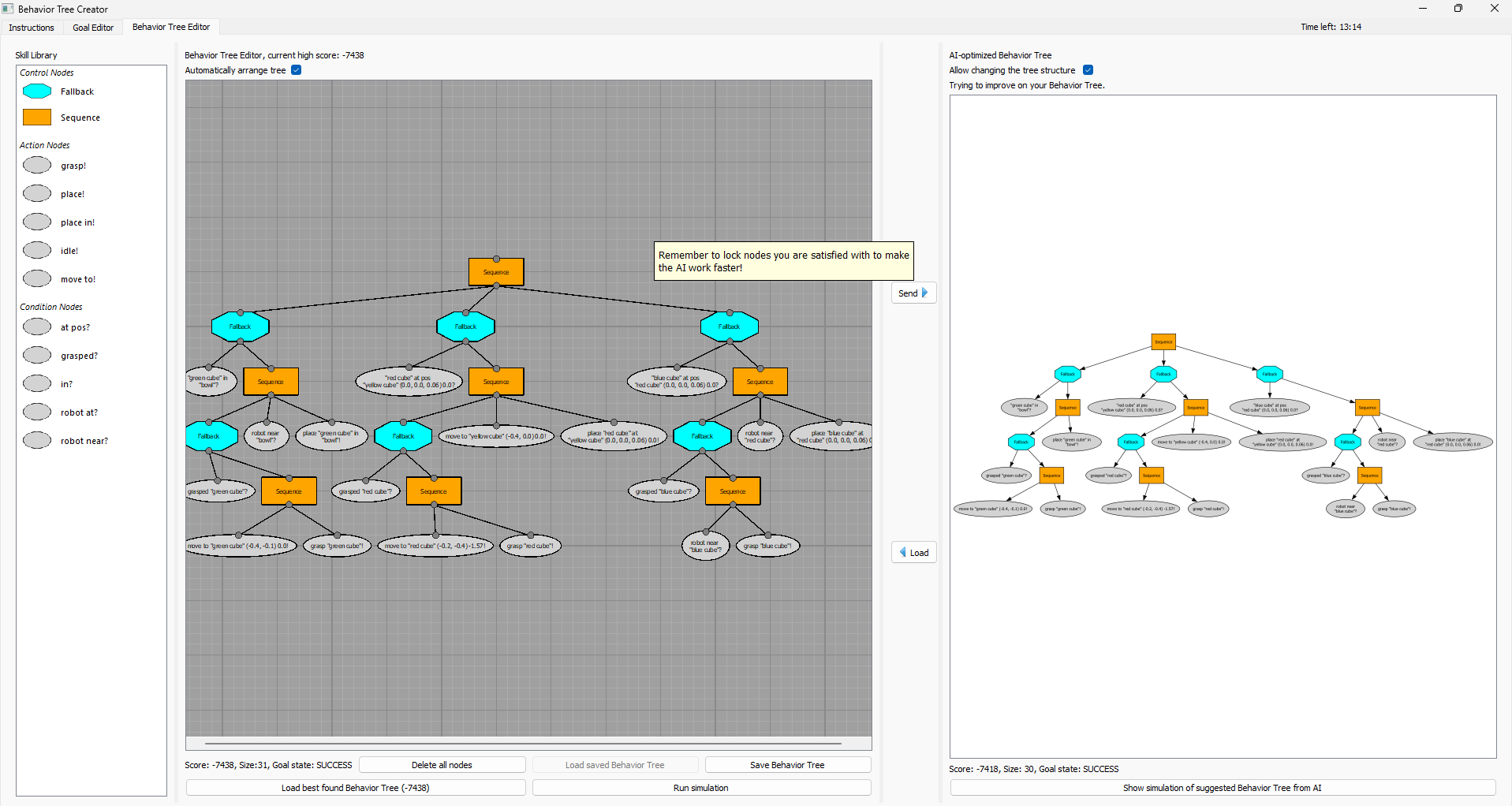}

\vspace{-0.0cm}
\caption{Screenshot of the Behavior Tree Editor tab of the GUI with the user's current tree to the left and the AI assistant's suggestion to the right. A hint to the user is shown in the textbox with a yellow background.}
\vspace{-.0cm}
\label{fig:tab3}
\end{figure*}
The third and final tab of the interface is the Behavior Tree Editor Tab, shown in Figure~\ref{fig:tab3}, where the user defines the BT policy to solve the given task. The editor widget and node list are mostly the same as in the Goal Editor Tab. One important difference is that the node list in this tab contains action nodes as well as the \emph{Fallback} node. The user also has buttons at the bottom of the tab to a) save the current tree, b) load a previously saved tree, c) delete all nodes, d) load the best tree found during the experiment, and e) run a simulation of the tree to evaluate it.
\subsection{AI Assistant}
\label{sec:assistant}
In all variants except MANUAL\_ONLY the user has access to an AI assistant that suggests improvements to the BT. The AI assistant combines elements from \cite{styrud_combining_2022}, \cite{styrud2024bebop} and \cite{styrud2025automatic} and some new additions into one cohesive system.
\par The AI assistant always works from a seed BT from the user as described below. Initially, this is a copy of the tree defined in the Goal Editor Tab, but the user can later update this by clicking the “Send” button and thereby using the tree currently in the editor widget as a seed to initiate the AI. After a new seed BT is sent to the AI, in all variants except \emph{NO\_PLANNER}, a PDDL-style planner is invoked to expand the tree. If the planner returns failure, the interface uses the method from \cite{styrud2025automatic} where an LLM is invoked to attempt to identify missing preconditions and insert them into the tree. This part is ablated in the \emph{NO\_LLM} variant. After adding the precondition(s), the planner is run again. These planner and LLM steps are iterated until either the planner succeeds or the LLM is unable to resolve the error.
\par The resulting tree after expanding with the planner and LLM is used to seed the GP or BO optimization algorithms (depending on user choice and variant). The optimization then continuously runs in the background while the user can continue working in the editor. As mentioned in Section~\ref{sec:gp} above, some constraints used in~\ours~were inspired by~\cite{mcclarron_effect_2016} with some differences, for example, allowing any control node type as the root. Another important constraint that we implemented is that two identical leaf nodes cannot be placed next to each other and that a control node may not have an identical control node as a parent, as these nodes would be redundant. We also use the “Conditional Precedence in Node Hierarchies” constraint as described in \cite{yang2025learning}, enforcing condition nodes to be placed before action when they are siblings.
\par Upon finding an improved tree, the user interface is updated with an image of the best tree found by the AI, as well as its score, size, and whether the tree successfully achieves the goal state or not. The user can then choose to load the suggested tree into the editor or run a simulation of the AI-suggested BT to evaluate its performance.
\par The user is also provided with the opportunity to lock one or more of the nodes in the tree sent as a seed to the AI. A locked node cannot be changed or removed by the AI. Other nodes can still be added. Locking nodes shrinks the search space and can significantly increase the speed of finding improved solutions.

\subsection{Simulation Window}
\begin{figure*}[hbtp]
\vspace{0.0cm}
\includegraphics[width=1.0\textwidth]{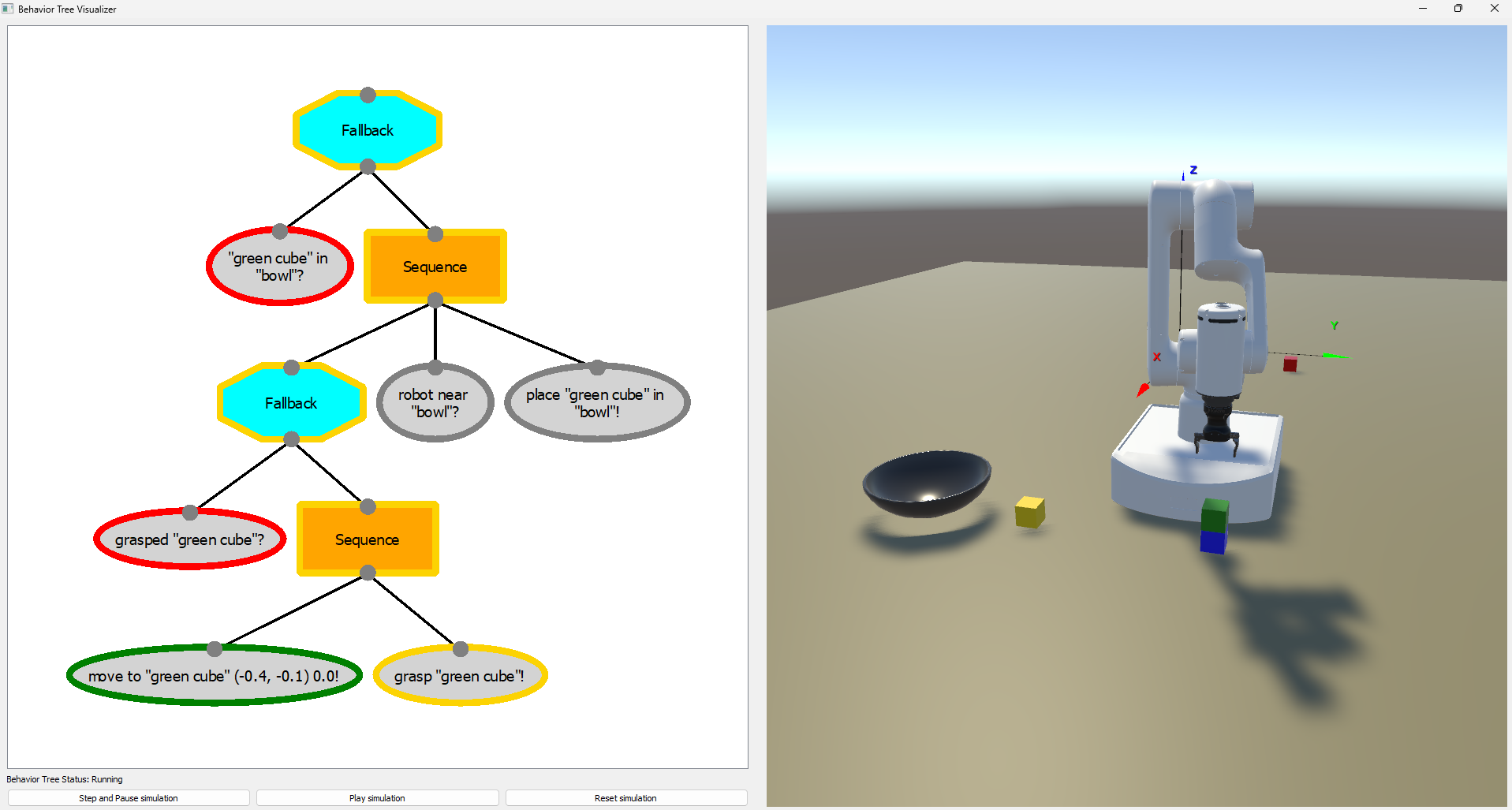}
\vspace{-0.0cm}
\caption{Screenshot of the Simulation Window of the GUI with the user's current tree to the left and the simulation on the right.}
\vspace{-.0cm}
\label{fig:simwindow}
\end{figure*}
When the user decides to run a simulation of a BT, a separate window opens up with the BT on the left and a 3D rendering of the simulation environment on the right, as shown in Figure~\ref{fig:simwindow}. The nodes of the BT in the Simulation Window each have a border where the color indicates the status of the node. Green means \emph{Success}, red means \emph{Failure}, yellow means \emph{Running} while grey means the node was not ticked. At the bottom of the simulation window are three buttons. “Reset simulation” returns all objects to their initial state and pauses the simulation. “Play simulation” resumes the simulation if it was paused. “Step and Pause simulation” executes one tick of the tree and then pauses the simulation.

\section{Task Design}
\label{section:task_design}
To evaluate the integration of an AI assistant in the GUI, four different task scenarios were designed in Unity. All scenarios include a 6-axis robot with a prismatic gripper mounted on a mobile platform with 3 degrees of freedom. The tasks were designed to be intuitive and easy to understand for novice users rather than being realistic examples of industrial robot tasks. The task design also needed to ensure that tasks were simple enough to be solvable within the allocated 15 minutes but not so simple that all users could trivially solve them. A typical tree size for a good solution for these tasks is around 15 nodes. We also needed to ensure that the tasks had solution BTs where structures and parameters of the tree had an impact, and where the planner and LLM did not solve the tasks immediately. All tasks are completely deterministic, as early tests indicated that it would be too difficult for users in the short time to construct trees handling random events. Finally, we made sure that each task was roughly equally difficult with the same number of subtasks, three, for each task.

\begin{itemize}
    \item The \emph{Demo} scenario is only used for a 5-minute learning session to get familiarized with the GUI before the user attempts to solve the other tasks. The scenario has three balls of different sizes and a goal area marked with a rectangle on the floor. The task goal is to move the red ball into the goal area.
    \item The \emph{Cubes and bowl} scenario has a bowl and four cubes of different colors. The task is to put the green cube in the bowl and to build a tower with the yellow cube at the bottom, followed by the red cube, and finally the blue cube on top. The scenario starts with the green cube on top of the blue cube, and the green cube must therefore be moved before the blue cube can be picked up.
    \item The \emph{Tableware} scenario has a table and a plate, knife, fork, and glass. The task is to set the table for a meal with the fork and knife on each side of the plate and the glass on the opposite side of the chair.
    \item The \emph{Trashpicking} scenario has three items of trash (paper, a banana, and a can) as well as three trash bins of different colors. The task is to sort the trash into the correct colored trash bin.
\end{itemize}
\subsection{Nodes}
Two control nodes are available to the users, \emph{Fallback} nodes and \emph{Sequence} nodes. None of the control nodes are set to have memory and thus tick the leftmost child every tick. The motivation is that with memory, tasks without random events can be solved using only a sequence of actions, but the intention of this experiment is to test the users' ability to build complete and reactive BTs.
The behavior nodes are designed to be at an abstraction level that is easy to understand in a short time but still leaves some room for the users to customize some parameters. No decorator nodes were included, as they were not necessary to solve the tasks, and it was essential to keep the overall interface simple.
\par Most behaviors have at least one parameter that can be chosen. The parameter \emph{target\_object}, when present, can be chosen from a list of movable objects in the scene. The \emph{relative\_object} is chosen from a list of all objects, static and movable. For the \emph{in?} condition and \emph{place in!} action, \emph{relative\_object} must be one of the container objects in the scene, such as a bowl, bin, or area. The \emph{offset} parameter is a tuple of three values (x,y,z) for objects or two values (x,y) for the robot base. The offset unit is meters. The \emph{angle} refers to rotation in radians around the z-axis.
\subsubsection{Condition nodes}
Five parameterized condition nodes are available to the user. For the tableware scenario, the \emph{in?} condition is excluded as it is not relevant for the task. 
\begin{itemize}
    \item The \emph{at pos?} condition checks if a \emph{target\_object} is at a position relative to a \emph{relative\_object} with an \emph{offset} and \emph{angle}. The allowed distance to this specified position is up to 2 cm and up to 0.2 radians.
    \item The \emph{grasped?} condition checks if a \emph{target\_object} is currently grasped by the robot.
    \item The \emph{in?} condition checks if a \emph{target\_object} currently inside a \emph{relative\_object}
    \item The \emph{robot at?} condition checks if the robot base is currently at the \emph{target\_object} with an \emph{offset} and \emph{angle}. The allowed distance to this specified position is up to 10 cm and up to 0.1 radians.
    \item The \emph{robot near?} condition checks if the robot base is within 0.9 m of the \emph{target\_object}.
\end{itemize}

\subsubsection{Action nodes}
Five parameterized action nodes are available to the user. For the tableware scenario, the \emph{place in!} action is excluded. In general, action nodes directly return \emph{Success} if the actions post-conditions are fulfilled even before execution. As an example, \emph{grasp!} returns \emph{Success} if the \emph{target\_object} is already grasped, but \emph{Failure} if the wrong object is grasped.
\begin{itemize}
    \item The \emph{grasp!} action moves the robot arm and grasps the \emph{target\_object}. 
    \item The \emph{place!} action moves the robot arm and places the \emph{target\_object} at a position relative to a \emph{relative\_object} with an \emph{offset} and \emph{angle}.
    \item The \emph{place in!} action moves the robot arm and places the \emph{target\_object} inside a \emph{relative\_object}
    \item The \emph{idle!} action does nothing and always returns \emph{Running}.
    \item The \emph{move to!} action moves the robot base to the \emph{target\_object} with an \emph{offset} and \emph{angle}
\end{itemize}

\subsection{Score}
After simulating an episode, each BT is given a score. The score is presented to the user as guidance for how good the BT is. It is also used as a target by the optimization algorithms of the AI assistant. As the convention is different in various research fields, the score could also be labeled, for example, “reward”, “cost”, or “fitness”, but in these experiments we exclusively refer to it as “score” to the users, as it is relatively neutral and simple to understand for novice users. The score is calculated after each BT tick and added up after the episode is completed. This indirectly rewards BTs that solve the tasks faster. In reality, all the terms of the score sum are negative penalties, and the objective therefore becomes to achieve a score as close to zero as possible.
\par The main part of the score is goal condition fulfillment. For each condition, a penalty is added proportional with a weight $\alpha$ to the distance of the target object from the goal position as well if the error is above some $\delta$. Similarly, a penalty is added proportional with another weight $\theta$ to the difference between the target object's orientation and the goal orientation, if the difference is above $\epsilon$. This goal part of the score is then given by Equation~\eqref{eq:score_goal}.
\begin{equation}
\label{eq:score_goal}
\begin{aligned}
\mathcal{S}_{goal} = - \sum_\mathcal{O} (&\alpha \cdot max(0, ||p - g_p|| - \delta) + \\ 
&\theta \cdot max(0, angle\_dist(r, g_r) - \epsilon))
\end{aligned}
\end{equation}
where $\mathcal{O}$ is the set of objects with specified goal poses, with current positions $p$, current orientations $r$, goal positions $g_p$, and goal orientations $g_r$. $||o - g||$ is then the distance error in meters, and $angle\_dist(r, g_r)$ is the angle error in radians.
\par Another significant part is related to the complexity of the tree. This penalizes large trees or trees where the execution logic is difficult to follow. Specifically, a penalty $\lambda$ is added for each node in the tree. Another penalty is added with the weight $\beta$ multiplied with the maximum depth of the tree. A penalty $\phi$ is added for BTs that return \emph{Failure} during the episode and another penalty $\tau$ for BTs that do not fail but exceed the maximum tick limit without returning~\emph{Success}. Further, a penalty $\psi$ is added for each action that fails during execution. To ensure a logical flow of ticks during execution, a penalty $\xi$ is added whenever an action is running that is further to the left in the tree than the most recently executed action. Finally, to ensure that objects are released after completion, a penalty $\eta$ is added for holding objects. This tree complexity score is then given by Equation~\eqref{eq:score_complexity}.
\begin{equation}\label{eq:score_complexity}
\begin{aligned}
\mathcal{S}_{tree} = \lambda L - \beta D - \tau T - \phi F - \xi R - \psi F_{actions} - \eta H
\end{aligned}
\end{equation}
where $L$ is the number of nodes and $D$ is the maximum depth. $T$ is 1 if the tree ended by timeout and 0 otherwise, and $F$ is 1 if the tree ended in Failure state and 0 otherwise. $F_{actions}$ is the number of actions that returned failure this tick. $R$ is 1 if the running action this tick is further to the left than the last previously run action and 0 otherwise. Finally, $H$ is 1 if the robot is holding an object and 0 otherwise.
\par The third part is a penalty for energy consumption. This is crudely approximated by the distance traveled by the robot, with the base movement being 10 times more costly than the movement of the robot arm. The energy part of the score is then given by Equation~\eqref{eq:score_complexity}.

\begin{equation}\label{eq:energy}
\begin{aligned}
\mathcal{S}_{energy} = \omega (A + 10B)
\end{aligned}
\end{equation}
where $\omega$ is a weight, $A$ is the distance traveled by the arm, and $B$ is the distance traveled by the base since the last tick.
The total score is finally given by Equation~\ref{eq:score_tot}.
\begin{equation}\label{eq:score_tot}
\begin{aligned}
\mathcal{S}_{total} = \mathcal{S}_{goal} + \mathcal{S}_{tree} + \mathcal{S}_{energy}
\end{aligned}
\end{equation}

\par Similar score functions have been used in previous work; the new terms added for this work were the penalty for actions running further left than the last tick and the energy penalty. If an episode is ended prematurely, for example, because the BT returns \emph{Failure}, the score from the final tick is multiplied with 200 minus the number of ticks executed. This is equivalent to each BT running exactly 200 ticks but saves the execution time of actually having to run all ticks.
\par The same weights for each of the terms are used for all the different task scenarios. The values of the weights were not extensively tuned but simply set manually to ensure that the score function fulfilled some basic properties, like small failing trees getting worse scores than a large tree that successfully solves the task. The weight values and the complete implementation are available on the project repository on GitHub\footnote{https://github.com/jstyrud/BETR-GUI}.

\subsection{Simulation}
The simulation environment was built with the Unity Engine, utilizing the integrated PhysX engine. The main focus was providing fast execution and a consistently deterministic experience to the user with clearly recognizable but not necessarily realistic visual elements. The dynamical and visual fidelity of the simulation was therefore considered secondary to speed and determinism for this study. A custom Google Protobuf-based API was built, to provide direct integration with the rest of the UI environment. Inspired by the Unity MLAgents \cite{juliani2020} and Gymnasium APIs \cite{towers2025gymnasiumstandardinterfacereinforcement}, it allows the simulation to be reset, stepped, and observed externally while providing control inputs for the agents. It also allows control over various simulation parameters, such as time scaling (relative to real time), timestep size, number of simulation steps per control input, and more.

The different experiment variants used in the study are organized as Unity scenes, with manually controlled physics simulation through the custom API. To improve simulation speed for the optimization-based variants, multiple agents can be loaded and executed in parallel, allowing for more overall throughput for the GP algorithm. For determinism, the physics scene is reloaded at the beginning of every execution to provide fully reinitialized physics bodies and remove all possible residual forces. The Articulation Body system provided by Unity was employed to model the robot and objects, as it proved more stable and less sensitive to floating-point indeterminism. To further enhance determinism, we did not consider friction between the internal components of the robot, and simulated rigid grasping of objects through the deployment of temporary joints, rather than using friction to interact between the target object and the robotic gripper.

\section{Experiments}
To study whether the system is effective and to what extent the various methods contribute, we conducted a user study testing the full system as well as method ablations and a manual-only variant that resembles the GUIs for BT creation that are available on the market today.

\subsection{GUI Variants}
\label{sec:variants}
A total of 6 different variants were implemented and tested in the experiments. 
\begin{itemize}
    \item The \emph{FULL} variant has all functionality included.
    \item The \emph{MANUAL\_ONLY} variant ablates all AI functionality, leaving the user to manually construct the BTs themselves. This is largely similar to existing GUIs like Groot~\cite{groot2_webpage} in capability.
    \item The \emph{NO\_BO} variant ablates BO from the AI's capabilities. If the user unchecks the “Allow changing the tree structure” checkbox in this variant, the GP algorithm continues running but is only allowed to mutate parameters.
    \item The \emph{NO\_GP} variant ablates GP from the AI's capabilities. In this variant, checking the “Allow changing the tree structure” allows the planner to run before the BO, while unchecking leaves only BO.
    \item The \emph{NO\_LLM} variant ablates LLM from the interface. This includes both the LLM assistant in the Goal Editor as well as the LLM used for error resolution in the Behavior Tree Editor.
    \item The \emph {NO\_PLANNER} variant ablates the planner functionality.
\end{itemize}

\subsection{Experimental Design}
We used a mixed groups design, where each participant interacted with a subset of three out of the six GUI variants. All participants always experienced both the \emph{MANUAL\_ONLY} and \emph{FULL} variants. In addition, they were randomly assigned to one of the four ablation variants (\emph{NO\_BO}, \emph{NO\_GP}, \emph{NO\_LLM}, \emph {NO\_PLANNER}). The subset of GUI variants that participants interacted with was therefore manipulated as a between-groups variable with 4 groups.
\begin{itemize}
    \item \emph{FULL}, \emph{MANUAL\_ONLY}, \emph{NO\_PLANNER} 
    \item \emph{FULL}, \emph{MANUAL\_ONLY}, \emph{NO\_GP}
    \item \emph{FULL}, \emph{MANUAL\_ONLY}, \emph{NO\_BO}
    \item \emph{FULL}, \emph{MANUAL\_ONLY}, \emph{NO\_LLM}
\end{itemize}

The task was manipulated as a within-groups variable with 3 groups (\emph{Cubes and bowl}, \emph{Tableware}, \emph{Trashpicking}). The trial order was also manipulated as a within-groups variable with 3 levels (first, second, and third). We randomized all the experiments in advance, ensuring that all ablations occurred the same number of times, and that the order in which tasks and variants occurred was also counterbalanced. The code for creating the randomization is available in the GitHub repository, and the resulting plan is uploaded to the OSF repository.


\subsection{Pilot Study}
The different tasks and GUI variants were piloted iteratively with 5 different participants, none of whom were part of the final study. Small updates were made after each pilot test to ensure that the tasks were neither too easy nor too difficult and could be reasonably completed within 15 minutes.

\subsection{Hypotheses}
We pre-registered our hypotheses and planned analyses (both confirmatory and exploratory) on the Open Science Framework (OSF)\footnote{\url{https://osf.io/ax5gb/overview}}.
\begin{enumerate}
\item Compared to all other variants (\emph{NO\_BO}, \emph{NO\_GP}, \emph{NO\_LLM}, \emph {NO\_PLANNER}, \emph{MANUAL\_ONLY}), \emph{FULL} will have: \begin{enumerate}
        \item Higher system usability
        \item Better task performance 
\end{enumerate} 

\item Compared to \emph{MANUAL\_ONLY}, all other ablation conditions (\emph{NO\_BO}, \emph{NO\_GP}, \emph{NO\_LLM}, \emph {NO\_PLANNER}) will have: \begin{enumerate}
       \item Higher system usability
       \item Better task performance
\end{enumerate} 

\item The \emph{FULL} variant used by a human will have better task performance compared to a baseline (\emph{NO\_HUMAN} with no human editing, full AI only) condition.
\end{enumerate}

\subsection{Participants}
The number of participants needed was determined \emph{a priori} using a simulated power analysis with $\alpha = .05$, suggesting 60 participants are sufficient to reach 80\% power, assuming small-medium effects\footnote{See supplementary materials for code to recreate the power analysis.}. 
\par We recruited 60 participants via flyers, mailing lists, social media, and word of mouth. After the experiments, we excluded one participant due to a one-time bug in the GUI, and recruited a new participant to redo the excluded experiment. In the end, the dataset includes data from 60 valid participants between the ages of 20 and 62 ($M_{age} = 29.7$, $SD = 8.9$).  Of the valid participants, 10 identified as female and 50 as male.  The distribution of participants across tasks and GUI variants can be seen in Table~\ref{tab:randomization}.

\begin{table}[tb]
\centering
\caption{Distribution of participants across tasks and GUI variants.}
\label{tab:randomization}
\begin{tabular}{rrrr}
  \hline
 & cubebowl & tableware & trashpicking \\ 
  \hline
FULL &  16 &  22 &  22 \\ 
  MANUAL\_ONLY &  22 &  20 &  18 \\ 
  NO\_BO &   7 &   4 &   4 \\ 
  NO\_GP &   7 &   5 &   3 \\ 
  NO\_LLM &   4 &   3 &   8 \\ 
  NO\_PLANNER &   4 &   6 &   5 \\ 
   \hline
\end{tabular}
\end{table}

The participants were primarily university students of engineering or computer science or professional software developers. To account for this, we asked participants about their familiarity with behavior trees, robots in general, programming in general, and programming robots using the following three questions: 

\begin{itemize}
    \item How familiar are you with \underline{\hspace{1cm}}? (1 = not at all, 5 = very)
    \item How often do you use or interact with \underline{\hspace{1cm}} (1 = never, 5 = often)
    \item How would you rate your level of expertise with \underline{\hspace{1cm}}? (1 = novice, 5 = expert)
\end{itemize}

The scores within each domain (behavior trees, robots in general, programming in general, and programming robots) were then averaged to form a general \emph{familiarity} score; see Table~\ref{tab:fam_means}.

\begin{table}[tb]
\caption{Participant's familiarity with behavior trees, robots in general, programming in general, and programming robots.}
\label{tab:fam_means}
\centering
\begin{tabular}{ll}
\hline
                       & Mean (SD) \\ \hline
Behavior Trees         & 2.13 (0.95)  \\
Robots in General      & 3.36 (1.06)  \\
Programming in General & 3.87 (0.81)   \\
Programming Robots     & 3.10 (1.08) \\ \hline
\end{tabular}
\end{table}

\subsection{Procedure}
\label{sec:procedure}
\begin{figure}[tpb!]
\centering
\includegraphics[width=0.45\textwidth]{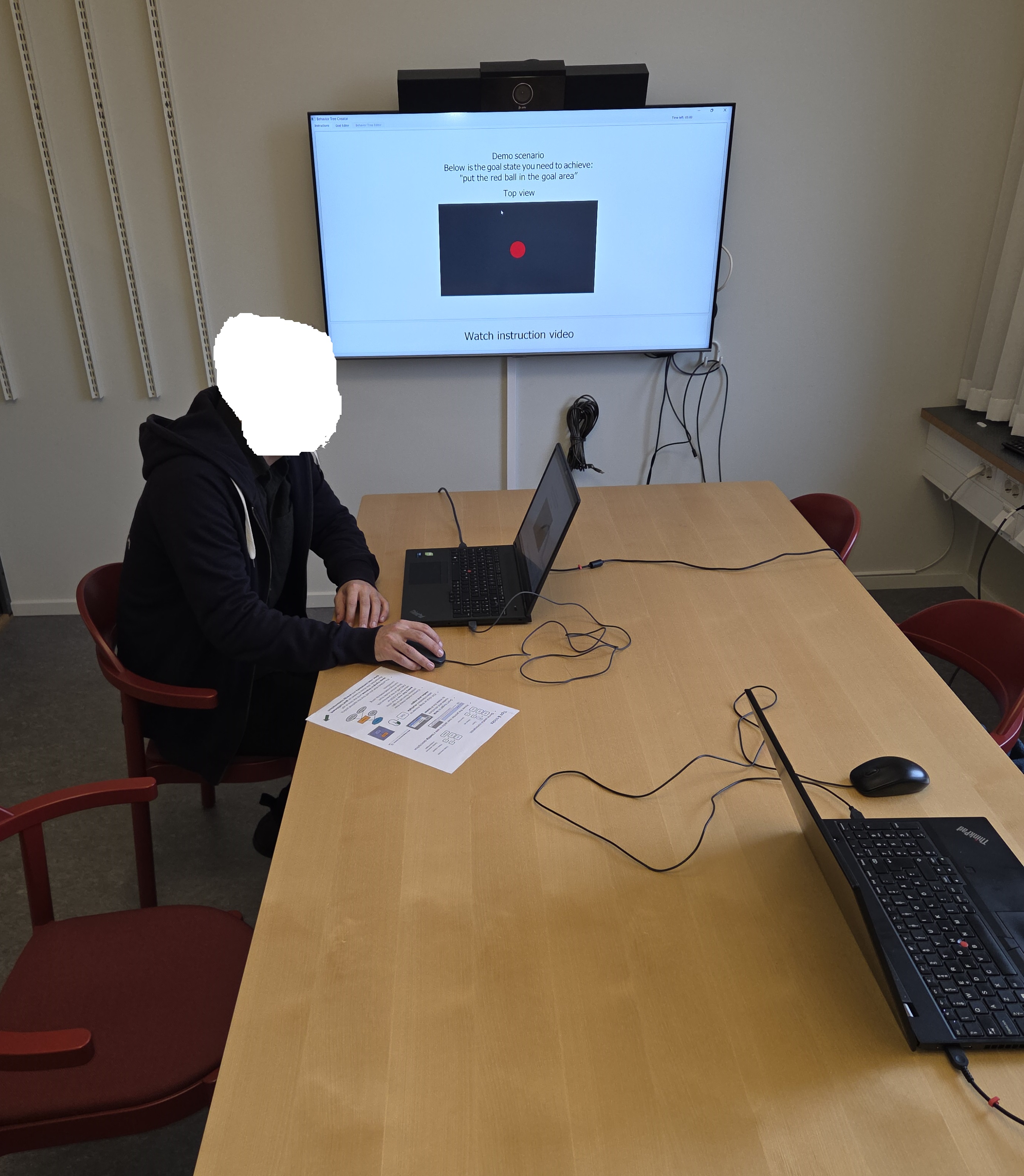}

\caption{Example setup with the participant using the laptop on the left with the screen duplicated onto the TV screen in the center. The experiment supervisor would be seated by the laptop on the right. Multiple rooms were used, but all had a similar layout.}
\vspace{-0.0cm}
\label{fig:experiment_setup}
\end{figure}
This study followed the ethical guidelines for Sweden, took approximately one hour for each participant, and was housed either at KTH in Stockholm or in ABB Robotics offices in Västerås. At the beginning of the experiment, participants were given an information sheet and asked to indicate their consent to participate. Participants were offered a gift card worth 100 SEK (approx. \$11 USD) as thanks for participating. 
\par For the experiments, the users were given a Lenovo ThinkPad laptop with an external mouse and an Intel(R) Core(TM) Ultra 9 185H 2.30 GHz CPU. To make sure there was no difference in the computing power for the learning methods, all participants used the same laptop. In general, the main computation bottleneck lies in the environment simulation, which is CPU-based. As seen in Figure~\ref{fig:experiment_setup}, the user's screen was duplicated onto a larger TV screen so that the experiment supervisor could follow along and ensure that the participant was following instructions and not, for example, looking for solutions on the internet.
\par After the consent form was signed, the participants were asked to answer some demographic questions as well as the questionnaire about their familiarity with robot programming, robots (in general), programming (in general), and behavior trees.
\par Once the questionnaire was signed, the participants watched a 5-minute instruction video introducing behavior trees and the GUI. The full video is available on the project repository. The participants were then given a one-page cheat sheet with a summary of the most important points of the video and some useful commands. They were allowed to keep the cheat sheet for the rest of the experiment. The participants were then able to spend 5 minutes using the \emph{MANUAL\_ONLY} GUI variant with the \emph{Demo} scenario to familiarize themselves with the workflow of the GUI.
\par Once the familiarization session was completed, the participants were given exactly 15 minutes for each of the three different scenarios, after which the GUI was automatically closed. The experiment timer starts counting down when the user first enters the goal editor tab, ensuring that all users get the exact same amount of time for each task. There were 6 available GUI variants as listed in Section~\ref{sec:variants}.  During the experiments, the GUI recorded actions taken and scores obtained by the participants, with timestamps.

After interacting with each GUI variant, participants were asked to complete the System Usability Scale~\cite{lewis2018system} on a 5-point scale from “\emph{1 - Strongly Disagree}'' to “\emph{5 — Strongly Agree}''. SUS scores were calculated according to~\cite{lewis2018system} which produces a score from 0-100. 


Finally, after completing all three tasks, the participants were asked to rank the three GUI variants they saw from most preferred to least preferred and were given a free-form text box to give any additional thoughts or feedback. 

\subsection{No human ablation experiments}
To test whether the human user added anything of value after the goal definition phase, we conducted ablation experiments where the AI assistant was allowed to run without any human doing any editing of the BTs. We refer to this as \emph{NO\_HUMAN}. For each of the 60 \emph{FULL} experiment runs, we identified the final goal set by the user, and the time spent to set the goal. The AI was then given 15 minutes (the full experiment time) minus the time spent to set the goal. For each experiment, we ran the AI five times with different seeds for a total of 300 runs.

\section{Results}
We performed mixed modeling using the \texttt{lme4} package in R~\cite{bates2015}. Mixed modeling allows us to account for individual differences between participants (e.g., familiarity with behavior trees) as well as the unequal distribution of GUI variants (i.e., that the \textit{FULL} and \textit{MANUAL} variants were seen more than the ablation variants). We mean-centered continuous variables. Models are reported using the guidelines set out in~\cite{bono2021report}, with pseudo $R^2$ used to estimate model fit. We compared models using corrected Akaike’s Information Criterion (AICc), where $\Delta AICc \leq 2$, we selected the simplest model~\cite{burnham2004multimodel}. For post-hoc comparisons, we used adjustments for multiple comparisons when appropriate~\cite{barnett2022multiple}. For simplicity, only results for the best-fitting models are reported here. Full analyses are available in the OSF repository.

\subsection{Normalization}
For readability, all task scores presented in the paper have been normalized. A score of 0 here represents the score of a minimal BT policy with two nodes that fails immediately and does nothing. The value 0 is, however, not the minimum possible score. In fact, there is no minimum score, as trees can be arbitrarily large and receive an arbitrarily large penalty for tree size. At the other end, 100 represents the maximum score achieved by any participant during the experiments. This is not an optimal score, as better scores were found in tests during development for each of the three tasks. The actual optimal achievable score is not trivial to compute. Scores displayed in the GUI during the experiments were not normalized and ranged typically from $-30000$ to $-4000$.

\subsection{Task Performance}
\begin{table}[tb]
\centering
\caption{Mean task score for each GUI variant.}
\label{tab:score_means}
\begin{tabular}{ll}
  \hline
GUI variant & Mean score (SD)  \\ 
  \hline
FULL & 91.14  (9.07) \\ 
NO\_BO & 90.38  (13.89) \\ 
NO\_GP & 82.01  (19.35) \\ 
NO\_PLANNER & 43.40  (21.99) \\ 
NO\_LLM & 39.91  (27.86) \\ 
MANUAL\_ONLY & 30.24  (35.08) \\ 
   \hline
\end{tabular}
\end{table}
We first compared the highest score achieved with each GUI variant; see Table~\ref{tab:score_means}. We constructed a Linear Mixed Model (LMM) with an identity link and random intercept using a maximum likelihood function. As fixed effects, we specified GUI variant, task, order, and familiarity with behavior trees to determine if and how each of these could predict task scores. For comparison, we also specified additional models testing for two- and three-way interactions between GUI variant, task, and order. This allowed us to determine if any differences between GUI variants remained consistent across trial order and tasks. 

\begin{figure}[tb]
    \centering
    \includegraphics[width=0.48\textwidth]{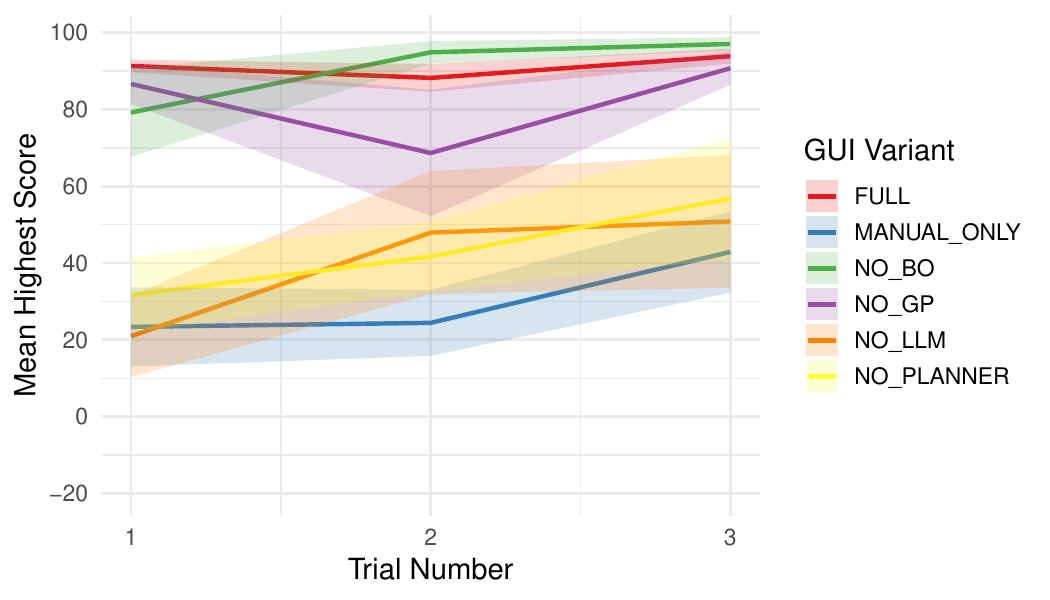}
    \caption{Mean task score across GUI variant and trial number.}
    \label{fig:score}
\end{figure}

The model containing only fixed effects (no interactions) provided the best fit for the data, $R^2 = 0.62$, see Figure~\ref{fig:score}. There were significant fixed effects of GUI variant, trial order, and behavior tree familiarity,  see Table~\ref{tab:score_fixef}. There was no effect of task type, indicating that the three different tasks were not significantly different from each other in terms of difficulty. 

\setlength{\tabcolsep}{4pt} 
\renewcommand{\arraystretch}{0.9}
\begin{table}[tb]
\centering
    \footnotesize
    \caption{Fixed and random effects for task score.}
    \label{tab:score_fixef}
\begin{tabular}{rcccccc}
  \hline
\textbf{Variable} & \textit{b} & \textit{se} & \textit{t} & \textit{F} & \textit{p} & 95\% CI \\ 
  \hline
(Intercept) & 79.16 & 5.58 & 14.18 & & $<$.001 & [68.2, 90.1] \\ 
GUI Variant &  &  &  & 295.17 & $<$.001 &  \\ 
Task &  &  &   & 2.28 & .320 &  \\ 
Trial Order & 6.88 & 1.99 & 3.45 & & $<$.001 & [3.0, 10.8] \\ 
BT Familiarity & 3.99 & 1.94 & 2.05 & & $<$.05 & [0.2, 7.8] \\ 
   \hline
\end{tabular}
\begin{itemize}
    \item[] Note: \textit{F}-values are reported where it is not possible to obtain a single \textit{b}-value estimate (i.e., for effects with >2 levels).
\end{itemize}
\end{table}

The fixed effects for trial number and familiarity with behavior trees were both positive, suggesting that participants' performance improved over trials and that participants with previous experience with behavior trees performed better. For the effect of the GUI variant, we conducted post-hoc pairwise comparisons between each GUI variant; see Table~\ref{tab:score_simple}. The results indicated that trials with the \emph{FULL} GUI variant, \emph{NO\_BO}, and \emph{NO\_GP} variants all had significantly higher scores than the \emph{NO\_LLM}, \emph{NO\_PLANNER}, and \emph{MANUAL\_ONLY} variants. There was no significant difference between the \emph{FULL}, \emph{NO\_BO}, and \emph{NO\_GP} variants nor between the \emph{NO\_LLM}, \emph{NO\_PLANNER}, and \emph{MANUAL\_ONLY} variants.

\begin{table}[tb]
\centering
    \footnotesize
    \caption{Pairwise comparisons for task score with GUI variants.}
    \label{tab:score_simple}
\begin{tabular}{lrrrr}
  \hline
\textbf{contrast} & \textit{b} & \textit{se} & \textit{t} & \textit{p} \\ 
  \hline
FULL - MANUAL & 61.05 & 4.13 & 14.79 & $<$.001 \\ 
FULL - NO\_BO & 1.71 & 6.76 & 0.25 & .999 \\ 
FULL - NO\_GP & 8.52 & 6.77 & 1.26 & .807 \\ 
FULL - NO\_LLM & 52.22 & 6.76 & 7.73 & $<$.001 \\ 
FULL - NO\_PLANNER & 47.73 & 6.72 & 7.10 & $<$.001 \\ 
NO\_BO - NO\_GP & 6.81 & 8.68 & 0.78 & .970 \\ 
NO\_BO - NO\_LLM & 50.51 & 8.74 & 5.78 & $<$.001 \\ 
NO\_BO - NO\_PLANNER & 46.02 & 8.71 & 5.28 & $<$.001 \\ 
NO\_GP - NO\_LLM & 43.70 & 8.77 & 4.98 & $<$.001 \\ 
NO\_GP - NO\_PLANNER & 39.21 & 8.71 & 4.50 & $<$.001 \\ 
NO\_LLM - NO\_PLANNER & -4.49 & 8.72 & -0.52 & .996 \\ 
MANUAL - NO\_BO & -59.34 & 6.73 & -8.82 & $<$.001 \\ 
MANUAL - NO\_GP & -52.54 & 6.73 & -7.80 & $<$.001 \\ 
MANUAL - NO\_LLM & -8.83 & 6.78 & -1.30 & .783 \\ 
MANUAL - NO\_PLANNER & -13.32 & 6.73 & -1.98 & .358 \\ 

   \hline
\multicolumn{5}{l}{{\footnotesize Results are averaged over the levels of task}}\\
\multicolumn{5}{l}{{\footnotesize To account for multiple comparisons, p-values were adjusted using tukey}}\\
\end{tabular}
\end{table}

\subsubsection{Comparison with \emph{NO\_HUMAN} condition}

Additionally, for the \textit{FULL} variant only, we tested whether the combination of human and AI outperformed a baseline of \emph{NO\_HUMAN}, where mean scores were 91.1 (9.07) and 88.06 (7.88), respectively. We specified an LMM with a random intercept using a maximum likelihood function and included fixed effects of score type (\emph{FULL} vs. \emph{NO\_HUMAN}) and task order. The model was significant, $R^2 = 0.04$, indicating that the \emph{FULL} condition had higher scores than \emph{NO\_HUMAN}, see Table~\ref{tab:lmm_score_type}. 

\begin{table}[tb]
\centering
\caption{LMM for the Effect of Score Type on Task Score}
\label{tab:lmm_score_type}
\begin{tabular}{lrrrrr}
\toprule
\textbf{Variable} & \textit{b} & \textit{se} & \textit{t} & \textit{p} & \textit{95\% CI} \\
\midrule
(Intercept) & 86.12 & 2.65 & 32.51 & 0.00 & [80.93, 91.31] \\
\emph{FULL} vs. \emph{NO\_HUMAN} & 3.08 & 0.89 & 3.46 & 0.00 & [1.33, 4.83] \\
Trial Number & 0.97 & 1.21 & 0.80 & 0.43 & [-1.40, 3.34] \\
\bottomrule
\end{tabular}
\end{table}

\subsection{System Usability Scale}
\begin{table}[tb]
\centering
\caption{Mean SUS score for each GUI variant}
\label{tab:sus_scores}
\begin{tabular}{lr}
  \hline
GUI variant & Mean (SD) \\ 
  \hline
FULL & 69.17  (16.77) \\ 
NO\_BO & 75.17  (19.19) \\ 
NO\_GP & 69.33  (9.14) \\ 
NO\_LLM & 51.83  (16.70) \\ 
NO\_PLANNER & 51.50  (23.64) \\ 
MANUAL\_ONLY & 59.33  (21.75) \\ 
   \hline
\end{tabular}
\end{table}
We used the SUS scores to analyze participants' subjective experience of the different GUI variants; see Table~\ref{tab:sus_scores}. As with task performance, we specified an LMM with a random intercept using a maximum likelihood function. We included fixed effects of GUI variant, task, order, and familiarity with behavior trees. We again specified additional two- and three-way models testing for interactions to account for any differences in the effect of GUI variant across tasks and trials. 

\begin{figure}[tb]
    \centering
    \includegraphics[width=0.48\textwidth]{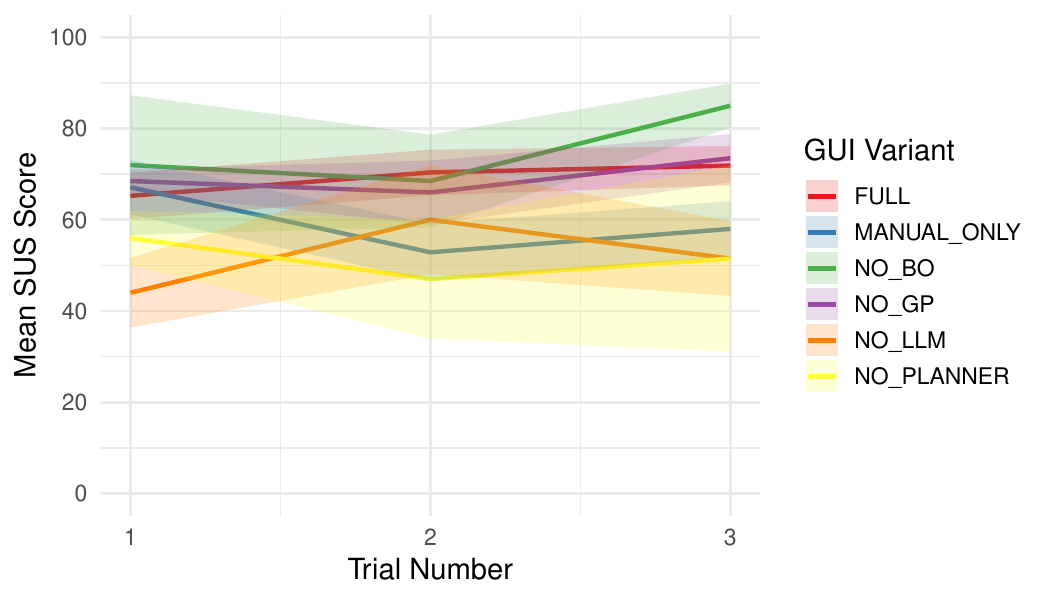}
    \caption{Mean SUS score across GUI variant and trial number.}
    \label{fig:sus_interact}
\end{figure}

The model containing only fixed effects (no interactions) provided the best fit for the data,  $R^2 = 0.21$, see Figure~\ref{fig:sus_interact}. There were significant fixed effects of GUI variant and familiarity with behavior trees, see Table~\ref{tab:score_fixef_sus}. There was no effect of task type or trial number, suggesting that perceived usability remained consistent across tasks and trials.

\begin{table}[tb]
    \centering
    \footnotesize
    \caption{Fixed and random effects for the SUS score.}
    \label{tab:score_fixef_sus}
    \begin{tabular}{lllllll}
      \toprule
  \textbf{Variable}   & \emph{b} & \emph{se} & \emph{t}  & \emph{F} & \emph{p} & \emph{95\% CIs} \\ \midrule
(Intercept) & 67.11 & 3.79 & 17.73 & & $<$.001  & [59.69, 74.53] \\ 
GUI Variant &  &  &  & 47.96 & $<$.001 &  \\ 
Task &  &  &   & 1.43 & .488 &   \\ 
Trial Order & 0.46 & 1.24 & 0.37 &  &  .713 & [-1.98, 2.89 ] \\ 
BT Familiarity & 5.49 & 1.98 & 2.77 & & $<$.01 & [1.60, 9.37] \\ 
   \hline
\end{tabular}
\begin{itemize}
    \item[] Note: \textit{F}-values are reported where it is not possible to obtain a single \textit{b}-value estimate (i.e., for effects with >2 levels).
\end{itemize}
\end{table}

There was a positive effect of familiarity with behavior trees on the SUS score, suggesting that as experience increased, all versions of the GUI were rated as more usable. Post hoc pairwise comparisons for the effect of GUI variant indicated the same pattern of results as task performance, see Table~\ref{tab:sus_simple}, with the \emph{FULL}, \emph{NO\_BO}, and \emph{NO\_GP} variants all outperforming the \emph{NO\_LLM}, \emph{NO\_PLANNER}, and \emph{MANUAL\_ONLY} variants. There was again no significant difference between the \emph{FULL}, \emph{NO\_BO}, and \emph{NO\_GP} variants nor between the \emph{NO\_LLM}, \emph{NO\_PLANNER}, and \emph{MANUAL\_ONLY} variants. 

\begin{table}[tb]
    \centering
    \footnotesize
    \caption{Simple effects for SUS score with GUI variant.}
    \label{tab:sus_simple}
    \begin{tabular}{lcccc}
  \hline
\textbf{contrast} & \textit{b} & \textit{se} & \textit{t}& \textit{p} \\ 
  \hline
FULL - MANUAL & 9.64 & 2.49 & 3.86 & .003 \\ 
FULL - NO\_BO & -7.31 & 4.33 & -1.69 & .540 \\ 
FULL - NO\_GP & -0.62 & 4.33 & -0.14 & .999 \\ 
FULL - NO\_LLM & 16.40 & 4.33 & 3.79 & .003 \\ 
FULL - NO\_PLANNER & 19.91 & 4.30 & 4.63 & 0.000 \\ 
MANUAL - NO\_BO & -16.95 & 4.30 & -3.94 & .002 \\ 
MANUAL - NO\_GP & -10.26 & 4.31 & -2.38 & .171 \\ 
MANUAL - NO\_LLM & 6.76 & 4.35 & 1.56 & .629 \\ 
MANUAL - NO\_PLANNER & 10.27 & 4.31 & 2.38 & .169 \\ 
NO\_BO - NO\_GP & 6.69 & 5.73 & 1.17 & .851 \\ 
NO\_BO - NO\_LLM & 23.72 & 5.78 & 4.10 & .001 \\ 
NO\_BO - NO\_PLANNER & 27.22 & 5.75 & 4.73 & 0.000 \\ 
NO\_GP - NO\_LLM & 17.02 & 5.81 & 2.93 & .045 \\ 
NO\_GP - NO\_PLANNER & 20.53 & 5.75 & 3.57 & .006 \\ 
NO\_LLM - NO\_PLANNER & 3.51 & 5.77 & 0.61 & .990 \\ 
   \hline
\multicolumn{5}{l}{{\footnotesize Results are averaged over the levels of task}}\\
\multicolumn{5}{l}{{\footnotesize To account for multiple comparisons, p-values were adjusted using tukey}}\\
\end{tabular}
\end{table}

\subsection{Rankings}

\begin{figure}[bt]
    \centering
    \includegraphics[width=0.48\textwidth]{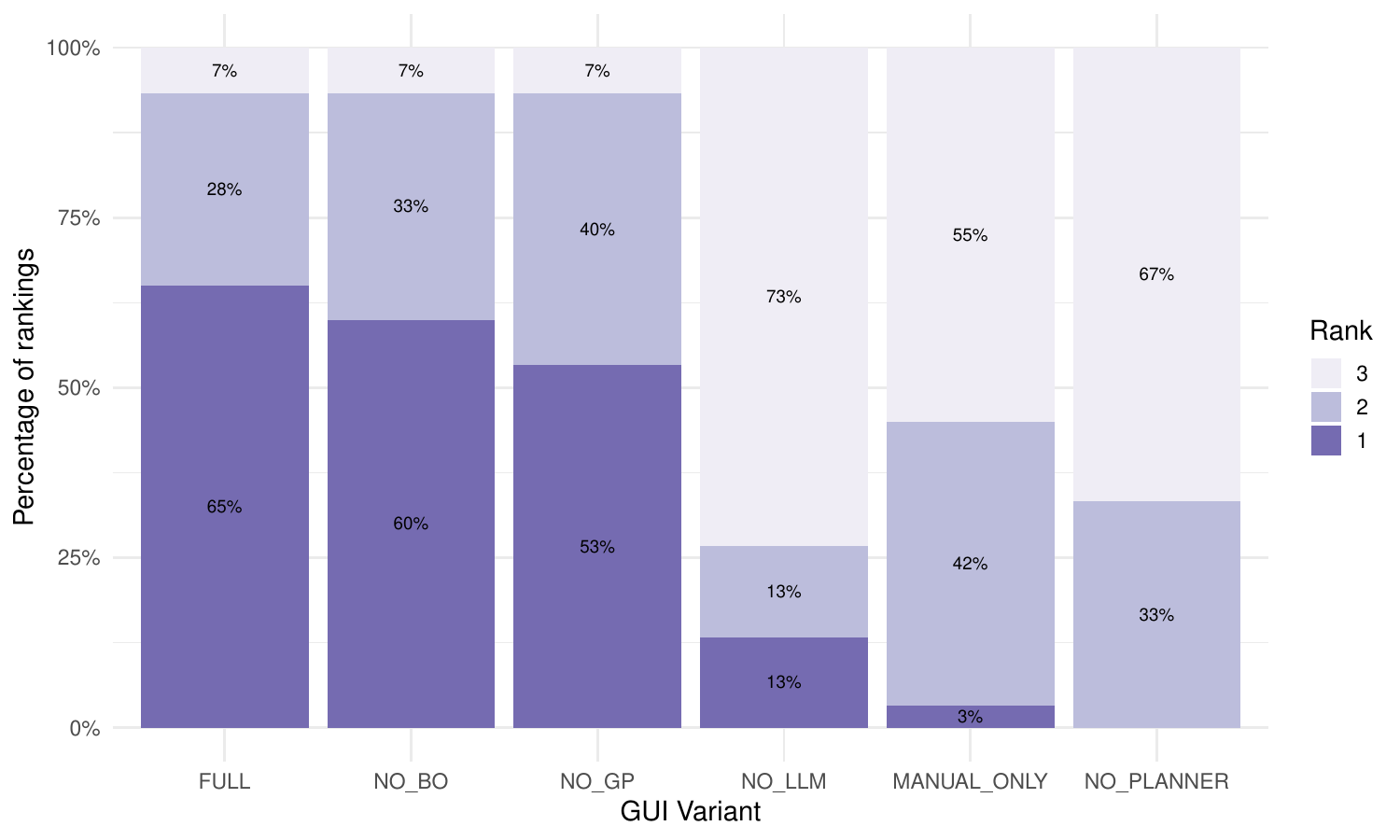}
    \caption{Proportion of rankings within each GUI variant.}
    \label{fig:gui_ranks}
\end{figure}
To analyze participants rankings of each GUI, we first calculated the percentage of times each GUI variant was selected for each rank; see Figure~\ref{fig:gui_ranks}.

We then constructed a cumulative link mixed model with a random intercept. The model was fit using maximum likelihood with a logit link function. Fixed effects included GUI variant and task. A likelihood ratio test showed a significant fixed effect of GUI variant, $\chi^2(5) = 404.2, p <.001$. 

Post-hoc pairwise comparisons confirm the same pattern of results, with the \emph{FULL}, \emph{NO\_BO}, and \emph{NO\_GP} variants all ranked significantly higher than the \emph{NO\_LLM}, \emph{NO\_PLANNER}, and \emph{MANUAL\_ONLY} variants. No other comparisons were significant.

\section{Discussion and Conclusions}
We present \ours, an interface for creating behavior trees that combines graphical programming with an AI assistant using LLMs, automated planning, Bayesian optimization, and genetic programming. We studied its effectiveness with an extensive user study with 60 participants, testing six different variants of~\ours~with specific functionality ablated. 
\par Our first and second hypotheses, that the \emph{FULL} variant would outperform all other variants, and that the \emph{MANUAL} variant would underperform compared to all other variants, were both partially supported. Primarily, the results indicate that users with access to \emph{FULL} AI assistant functionality achieve significantly higher scores in the experiments than the \emph{MANUAL\_ONLY} variant without the AI assistant. This stipulates the importance of including interaction with AI assistants in graphical user interfaces. The two ablations removing learning parts of the AI assistant, \emph{NO\_BO} and \emph{NO\_GP}, resulted in lower average scores in the experiments than \emph{FULL}, but the difference was not significant. There are several possible reasons for this. One possible reason is that the users didn't understand how to make full use of the AI to guide the learning algorithms. For example, in experiments where the participants had access to an AI assistant, only in 74/120 experiments was the locking node functionality  ever used, and only in 72/120 were nodes locked for more than 30 seconds, despite pop-up hints in the GUI suggesting that they should be used. 
Another possible reason is that in the benchmarking tasks, the planner typically solved the majority of the task, and the learning algorithms were left to mostly fine-tune the solutions, which didn't have as large an effect on the score. To test this, a new user study would have to be done with either different tasks that put more emphasis on the learning part, or with more participants to be able to separate variants that are closer in performance. Also, learning algorithms do not perform best in very short experiments like these, as they tend to need more time. In real settings, a user could leave the optimization running overnight, for example.
\par Results also showed that the participants performance improved significantly over the course of the experiment as they gained experience. Many users also explicitly told us that they found it difficult in the beginning. Besides learning the GUI, many users also had to learn how behavior trees worked, and participants with previous experience performed significantly better in the experiments. In the form, users wrote comments like \textit{“Hard to grasp how BT works”} and \textit{“Took me a while to get the fallback nodes and how to use them.”}. There was no significant difference in performance on the three benchmark tasks, indicating that they were roughly equal in difficulty, which was also the intention during design.
\par The ablations \emph{NO\_LLM} and \emph{NO\_PLANNER} did not get significantly higher scores than the \emph{MANUAL\_ONLY} variant. With these two ablations, the initial suggestions from the AI assistant will not solve even a single subtask, giving very low scores. While the AI assistant in these variants is still potentially useful if the user is able to guide it by sending suggested BTs. In the experiment, many users either were unable to solve the basic structure by themselves in time, or simply determined that the AI was of no use and elected not to make use of it. This is also suggested by comments that the users wrote, like, \textit{“When AI gives a bad advice, you don't trust it again. Rather fix it yourself than fix the broken AI solution.”}, \textit{“The AI in GUI 2 confused me more than it helped.”}, and \textit{“I did NOT trust the last AI, it seemed like it was out to get me. :'(”}. While the differences are not statistically significant, it's interesting to note that \emph{NO\_LLM} and \emph{NO\_PLANNER} scored lower on the SUS while getting higher mean scores on task performance, suggesting that there were issues with trusting the AI. Indeed, even in some experiments with \emph{FULL} functionality, some users chose not to load the AI suggestions even though it solved the task while stating that they thought they could do better themselves, and thereafter failed to find any solution manually. This suggests that building trust with the user could be important for overall system performance.
\par Our final hypothesis, that the \emph{FULL} variant would outperform the \emph{NO\_HUMAN} ablation experiments, was fully supported. The results showed that the AI running by itself without any help from the human user performed significantly worse when limited to the same time. This shows that even with this rather extensive AI assistant, the human still plays an important role in the behavior tree design effort.

\section{Future work}
The main limitation of this work is that the benchmark tasks, out of necessity, are highly simplified compared to actual robot applications. An important future work would be to perform a study with realistic, complex tasks. In order for such tasks to be possible, it would also be necessary to give the users considerably more time for each experiment as well as more extensive training before the experiments, both regarding behavior trees and the GUI itself and its capabilities.
\par There are also numerous improvements that could be done to the GUI. Some of the more interesting ideas for this include:
\begin{enumerate}
    \item Planning directly in the editor, expanding nodes directly when placed.
    \item With a larger skills pool needed for more complex tasks, let the user select which skills should be enabled for the AI. This selection could also be done with another AI.
    \item Auto-complete functionality where an LLM suggests additions to the tree without testing in simulation.
    \item Use a completely LLM-generated tree, for example, using a method from previous work~\cite{izzo2024btgenbot, mower2024ros, lykov2023llm} and insert it into the GP population.
    \item A prompt functionality to send natural language suggestions to the AI assistant also in the editing stage, such as “No, move the red cube first.”
\end{enumerate}
\par Important to note is that the usefulness of the various methods in a system such as~\ours~will not only be dependent on the methods themselves, but also on the capability of the other methods in the system. For example, if a planner is capable of optimally solving the task at hand in an acceptable time, there is no need for a learning algorithm. Therefore, as each method of~\ours~is developed further, future work must continuously reevaluate the performance, the roles, and the relative importance of the various components.
\par Finally, it could be important to study what can be done to increase trust in the AI assistant in terms of, for example, the assistant's functionality, GUI design, and user training.

\bibliographystyle{IEEEtran}
\bibliography{IEEEabrv, biblio, references, llmbiblio, thesisbib}

\end{document}